\documentclass[11pt]{article}
\usepackage{enumitem}
\usepackage{algorithm}
\usepackage{multirow}
\usepackage{makecell}
\usepackage{fullpage} % Reasonable margins
\usepackage{booktabs}
\usepackage{graphicx}
\usepackage{natbib}
\usepackage{amsmath}
\usepackage{amssymb}
\usepackage{styles/cme-math}

\if@nosubcaption
   \usepackage{subfigure}
\else
    \usepackage{subcaption}
\fi

\AtBeginDocument{%
  }
    
\begin{document}

\title{Outbound Modeling for Inventory Management}

\author{
  Riccardo Savorgnan\textsuperscript{1}\thanks{Both authors contributed equally to this research.} \and
  Udaya Ghai\textsuperscript{2}\footnotemark[1] \and
  Carson Eisenach\textsuperscript{1} \and
  Dean Foster\textsuperscript{1}
}

\date{
  \textsuperscript{1}Amazon – SCOT Inbound Systems, New York City, NY, USA\\
  \textsuperscript{2}Amazon AWS – NeuroSymbolic AI, New York City, NY, USA\\[1ex]
  \texttt{\{savorgr, ughai,ceisen, foster\}@amazon.com}
}

\maketitle
\begin{abstract}
We study the problem of forecasting the number of units fulfilled (or ``drained'') from each inventory warehouse to meet customer demand, along with the associated outbound shipping costs. The actual drain and shipping costs are determined by complex production systems that manage the planning and execution of customers' orders fulfillment, i.e. from where and how to ship a unit to be delivered to a customer. Accurately modeling these processes is critical for regional inventory planning, especially when using Reinforcement Learning (RL) to develop control policies. For the RL usecase, a drain model is incorporated into a simulator to produce long rollouts, which we desire to be differentiable. While simulating the calls to the internal software systems can be used to recover this transition, they are non-differentiable and too slow and costly to run within an RL training environment. Accordingly, we frame this as a probabilistic forecasting problem, modeling the joint distribution of outbound drain and shipping costs across all warehouses at each time period, conditioned on inventory positions and exogenous customer demand. To ensure robustness in an RL environment, the model must handle out-of-distribution scenarios that arise from off-policy trajectories. We propose a validation scheme that leverages production systems to evaluate the drain model on counterfactual inventory states induced by RL policies. Preliminary results demonstrate the model's accuracy within the in-distribution setting.
\end{abstract}
\section{Introduction}\label{sec:intro}

Today, reinforcement learning (RL) is used to optimize inventory management systems \citep{Madeka2022} that must decide order quantities to stock products used to fulfill customer demand. \citep{Madeka2022} modeled the stock buying problem as a periodic review inventory control where at each review period, a single action (representing the buying quantity for the whole country) is taken for each product and the state tracks the total inventory, making the dynamics just accounting of what is received and sold. By assuming the retailers action do no affect demand, product level economics, lead-times and other relevant covariates, \citep{Madeka2022} fit the problem into the framework of an Exogenous Interactive Decision Process (ExoIDP, where exogeneity refers to the independence of a subset of covariates in respect of the agent's actions). Unlike worst-case RL, which suffer from exponential sample complexities, backtesting an ExoIDP can be analyzed directly with concentration bounds, and hence reduces to supervised learning. This allows us to reliably backtest on historical data in a gym by replaying the historical exogenous data for each product and tracking the counterfactual trajectories induced by a new policy.

More recently, RL is being proposed to tackle buying and placement supply chain problems which require finer-grained inventory state tracking. In particular, RL has been proposed to solve the the multi-echelon \citep{clark1960optimal} inventory control problem, wherein the action-space is expanded to be an order quantity for each warehouse (sometimes we will refer to these as nodes), or at a warehouse up in the serial line to then be re-routed downstream to warehouses dedicated to demand fulfillment. In such settings, the distribution of inventory across different nodes is necessary for optimal decision making and simulation of the dynamics. While \cite{Madeka2022} could afford to represent the state with total inventory, in order to properly capture the impact of such policies, we must represent the inventory state of each node\footnote{Possibly, other coarser granularity might work in some instances.}. Understanding the evolution of this state after a single period is far more complicated since there are \emph{exponentially} many ways for demand to be fulfilled. Furthermore, each possible configuration of inventory drain yields different fulfillment costs. The way this inventory is fulfilled can be complicated and dependent on many internal systems, along with many other external factors. To maintain the reduction to supervised learning, we would be required to be able to execute production systems with \emph{perfect fidelity}. 

While executing production systems may provide a good backtest for a regional inventory control policy, in order to {\it learn} a policy we need access to a high-speed, accurate, simulator where we evaluate and potentially differentiate through billions of rollouts -- a task for which our production systems are too high-latency and costly to run. Instead, we propose a {\it Drain Model} which will emulate the drain and shipping cost induced by an inventory configuration and incoming arrivals over some specified time period. We consider this as a forecasting problem and train a deep learning model for our task using historic, observational data.

One possible challenge with employing observational data is that it is derived from a distribution induced by historical buying and placement policies. A new regional inventory control policy would produce different state-action trajectories than the historical data distribution. A simulator (gym) for learning a regional inventory control policy must reliably track the real-world, necessitating off-policy coverage, a common requirement in RL. To illustrate this, consider the following scenarios:

\noindent
\textbf{Scenario A (Current Policy):} Inventory is placed unevenly, with 10 units in NYC and 0 units in LA. When 4 units of demand arise in LA, fulfillment must occur from NYC. This results in 6 units remaining in NYC, 0 units in LA, and high shipping costs due to cross-regional fulfillment.

\noindent
\textbf{Scenario B (Proposed Policy):} Inventory is distributed more evenly, with 5 units in both NYC and LA. When the same 4 units of demand arise in LA, they can be fulfilled locally. This results in 5 units remaining in NYC, 1 unit in LA, and low shipping costs due to local fulfillment.

\noindent
The following table summarizes these two scenarios:

\begin{table}[h!]
\centering
\caption{Comparison of current and proposed inventory placement policies.}
\small % Reduce font size of the table
\renewcommand{\arraystretch}{1.2} % Adjust row spacing
\setlength{\tabcolsep}{6pt}       % Adjust column spacing
\begin{tabular}{lcccc}
\toprule
\textbf{Scenario}       & \makecell{\textbf{Initial Inventory} \\  \textbf{(NYC/LA)}}  & \makecell{\textbf{Demand} \\ \textbf{(LA)}} & \makecell{\textbf{Fulfillment} \\ \textbf{(From)}} & \makecell{\textbf{Remaining Inventory} \\ \textbf{(NYC/LA)}} \\
\midrule
\textbf{A: Current Policy} & 10 / 0                            & 4                    & NYC                         & 6 / 0                                  \\
\textbf{B: Proposed Policy} & 5 / 5                             & 4                    & LA                          & 5 / 1                                  \\
\bottomrule
\end{tabular}
\label{tab:placement_policies}
\end{table}

The critical observation is that historical data only captures outcomes from Scenario A, where inventory is concentrated in NYC. The data distribution lacks examples of how the system would behave under alternative policies like Scenario B, where inventory is spread across warehouses. Consequently, a model trained solely on historical data might fail to predict outcomes for policies that deviate from historical patterns. To address this, our approach involves evaluating the drain model on distributions induced by alternative policies. This requires an oracle capable of generating off-policy validation data. Additionally, fine-tuning the drain model on data produced by the oracle may be necessary to ensure reliable simulation performance.

\section{Additional Background}\label{sec:related}

\paragraph{Fulfillment}
Fulfillment decisions are taken by a planning system -for brevity F- that optimizes the fulfillment cost of {\it sets} of orders, while respecting the original delivery dates that were shown to customers. While it would be ideal to use F as part of a learning framework for inventory control and placement policies, doing so is impractical due to the high latency of the calls and technical limitations when generating counterfactual actions. In this paper we thus set off with the objective of providing an accurate customer order fulfillment emulator, capable of simulating counterfactual fulfillment actions of the system F. 

\paragraph{Outbound back-testing oracles}
We need to validate our system dynamics model on states that are unseen in on-policy data trajectories. The actual system F is the desired instrument for this, as we can call it to generate counterfactuals on an arbitrary inventory state and compare it to the output of the model. To do so, two components are necessary: a) the ability to track per-product inventory along a simulated trajectory and b) customer responses to different products availability.  It will thus be possible to validate and collect data for an outbound model with counterfactual state/action trajectories. We propose an algorithm for replaying instances of customers viewing product webpages, which represent potential customer interest in buying an item, simulating customer responses to the counterfactual inventory availability and calling the production F algorithm to generate counterfactual inventory drain trajectories. This service could be used to validate an outbound model under ``off-policy'' state distributions; we discuss this in detail in \ref{subsec:conditional_conversion}.

\paragraph{Customer Demand Correction}
Delivery dates are influenced by local inventory availability. A customer who is offered a faster delivery has a different probability of buying an item. Furthermore, the delivery date also impacts the ship options available to a customer at checkout, and thus the probability of selecting different options. When changing local inventory availability, we may wish to adjust historical demands and ship options selected by the customer. We utilize a model for correcting demand under different local availability inventory configuration, based on modeling conversions from customer arrivals (i.e. webpage visits or glance views) to an order and ship option as distributed according to a multinomial distribution. We discuss the model in details in \ref{sec:gvconversion} and provide an extension in \ref{sec:backtest-outbound}.

\section{Mathematical Formulation}\label{sec:formulation}

\subsection{Notation}\label{sec:notation}

Throughout the rest of this paper, we will denote matrices by bold uppercase characters. For a set $S$, we denote the cardinality of that set by $|S|$. The notation $[N]$ denotes the sequence of natural numbers through $N$ (i.e. $\{1,...,N\}$). For a matrix $\Mb$, we denote the element in the $i$-th row and $j$-th column of the matrix as $\Mb_{i,j}$. We denote by $\RR_{\geq 0}$ the nonnegative reals and $\RR_+$ the positive reals. Similarly we denote $\ZZ_{\geq 0}$ and $\ZZ_+$ for the nonnegative and positive integers, respectively. Denote by $(\cdot)^+$ the positive part operator. We use $\Delta^d \subseteq \RR_{\geq 0}^{d+1}$ to represent the $d$ simplex.

\subsection{Outbound process}
\label{sec:outbound-overview}

In this section we describe the process of how outbound is determined, starting from a customer's interest in a product. Denote by $\cA$ the set of products managed by the retailer, $\cF$ the set of warehouses from which the retailer can fulfill orders and $\cZ$ the set of customer regions (for example a region can be defined by grouping all addresses sharing the first 2 digits of the zip code, also called Zip2s). First, the customer arrives at the detail page -i.e. the webpage- for a product $i \in \cA$. They are then shown a {\it promise} $p$ of how quickly the product can be delivered to them. There are multiple possible promises -- including one for {\it product is unavailable} -- and we denote the set of all promises by $\cS$. After viewing the promise, the customer decides whether or not to purchase the product. When purchasing the product, the customer selects a shipping speed option (for example, next-day or two-day) $o$ that is no faster than the original promise shown. The set of all ship speeds is denoted by $\cO$, and by convention we always include a ship-speed option that corresponds to {\it no order} being placed. Once the order is placed, the fulfillment policy F determines which warehouse to fulfill from. 

\subsubsection*{Exogenous and control processes}

Having described how the outbound process works, we define the processes that we wish to model and which determine the transition dynamics. We index time series processes by $t \in \ZZ_{\geq 0}$ and all time series are discretized to the same granularity (e.g. weekly). The time step $t$ corresponds to the time interval $[t, t+1)$.
\paragraph{Glance Views} A {\it glance view} consists of the associated product, the region of the customer, a promise shown, a ship option (SO) selected at checkout, and a quantity in $\RR_{\geq 0}$ ordered. Formally, a glance view $v$ belongs to  $\cV := \cA \times \cZ \times \cS \times \cO \times \RR_{\geq 0}$. We will denote by $H_Z$ the mapping from $v$ to its region $z \in \cZ$, and $H_A$, $H_S$, $H_O$ analogously, with $H_Q(v)$ producing the order quantity. At each time period $t$, and for each product $i$, there is a sequence of glance views $G^i_t = (v^i_{t,\tau})_{\tau \in \ZZ_{+}}$,where each glance view is associated with item $i$, which are indexed in increasing order of time $\tau$ that customers arrive at the detail page (each individual glance view typically corresponds to a different costumer). Next, we denote the total regional glance views for product $i$ at time $t$ as $g^i_t \in \RR^{|\cZ|}$ where $g^i_t := (g_t^{i,1},\dots,g_t^{i,|\cZ|})$ and 
\[
g_t^{i,z} := | \{ v \in G^i_t : H_Z(v) = z \} |.
\]

\paragraph{Active warehouses} The set of active warehouses changes over time, as new ones are built. We denote with $\tilde{\cF}_t \subseteq \cF$ the set of warehouses that are active at time $t$.

\paragraph{Inventory} We denote the inventory at a warehouse $f$ of product $i$ at the end of period $t$ as $I_t^{i,f} \in \RR_{\geq 0}$ and the vector of inventory for a product $i$ at time $t$ as $I^i_{t} \in \RR_{\geq 0}^{|\cF|}$ where $I^i_{t} := (I_t^{i,1},\dots,I_t^{i,|\cF|})$. 

\paragraph{Outbound units} We denote the units outbounded of product $i$ from a warehouse $f$ during time $t$ as $o^{i,f}_{t} \in \RR_{\geq 0}$. The vector of outbound for a product $i$ at time $t$ is defined as $o^i_{t} \in \RR_{\geq 0}^{|\cF|}$ where $o^i_{t} := (o_t^{i,1},\dots,o_t^{i,|\cF|})$.

\paragraph{Shipping cost} We denote the total shipping cost from a warehouse $f$ for product $i$ at time $t$ as $c^{i,f}_t$, and the vector of shipping costs for item $i$ at time $t$ as $c^i_{t} \in \RR_{\geq 0}^{|\cF|}$ where $c^i_{t} := (c_t^{i,1},\dots,c_t^{i,|\cF|})$. Note that the warehouse-level costs are the sum of the unit level costs for all units outbounded from that warehouse.  We discuss several ways to perform this unit-level accounting in \ref{sec:cost-accounting}.

\paragraph{Stowed units} We denote the units stowed (i.e. received and available for customer fulfillment) of product $i$ from a warehouse $f$ during time $t$ as $a^{i,f}_{t} \in \RR_{\geq 0}$. The vector of stowed units for a product $i$ at time $t$ is defined as $a^i_{t} \in \RR_{\geq 0}^{|\cF|}$ where $a^i_{t} := (a_t^{i,1},\dots,a_t^{i,|\cF|})$.

\paragraph{Historical covariates} Additionally, we may have other historical covariates of interest (such as holiday indicators, etc) that we may wish to include as part of the outbound modeling. We denote these as $y_{t}^i$.

\subsection{Formulation as a Forecasting Problem}
\label{sec:forecast-single-section}

Outbound quantities and shipping costs depend on glanceviews (customer arrivals) and inventory position, but the arrivals and conversions are stochastic. We thus cast this as a probabilistic forecasting task, seeking
\[
p(o^i_t, c^i_t \mid H^i_t, \theta) 
~=~
p(o^i_t \mid H^i_t, \theta)\,\,
p(c^i_t \mid o^i_t, H^i_t, \theta),
\]
where \(\theta\) are learnable model parameters.

\section{Proposed ML-Based Forecasting Approach}\label{sec:model}
\subsection{Outbound Distribution}
Similarly to \cite{Wang_2027}, because many product-warehouse-time combinations have small or zero outbound, we allocate explicit probability mass to those values, while accommodating unbounded large outbounds via the quantile-based tail. A fully discrete model would require enormous support, while a pure quantile-based approach often struggles with calibration when much of the data is zero or near zero. Consequently, we introduce a hybrid \emph{discrete-plus-quantile} model for 
\(p(o^i_t \mid H^i_t, \theta)\). 
Specifically, we define discrete probabilities 
\(p_{\text{disc}, k}(H^i_t, \theta)\) 
for integer values 
\(k = 0,1,\dots,n_d - 1\), 
with the remaining probability mass collected in 
\[
p_{\text{disc}, n_d}(H^i_t, \theta) 
~:=~ 
p\bigl(o^i_t \ge n_d\mid H^i_t, \theta\bigr).
\]
A learned quantile-based CDF 
\(F_q(k) := F_{\text{quant}}(k \mid H^i_t, \theta)\) 
then refines how this tail mass is distributed. Formally,

\begin{equation}
\label{eq:piecewise-outbound-tight}
p(o^i_t = k \mid H^i_t, \theta)
=
\begin{cases}
p_{d, k}\bigl(H^i_t, \theta\bigr), 
& k < n_d,\\
p_{d, n_d}\bigl(H^i_t, \theta\bigr)\,
\Bigl[
F_q(k) - F_q(k-1)
\Bigr], 
& k \ge n_d.
\end{cases}
\end{equation}

\subsection{Cost Distribution.}
Shipping costs do not exhibit the same degree of sparsity due to the conditioning on the outbound quantities labels, i.e. $p(c^i_t = 0 \mid o^i_t = 0, H^i_t, \theta) = 1$ the zero shipping cost can be predicted perfectly when we condition on zero outbound units, so we adopt standard quantile forecasting methods~\citep{Wen2017,Eisenach2020}, predicting \(n_q\) quantiles of \(c^i_t\) conditional on the realized outbound \(o^i_t\). Several techniques exist for interpolating these quantiles and drawing samples~\citep{Quenneville-Belair2023}.

\subsection{Loss function}
\label{sec:lossfunction}
The model is trained on a linear combination of $5$ losses. First, for the discrete outbound prediction, we use cross entropy loss. Then for the quantile predictions of the model, both in cost and in outbound, we use a combination of quantile loss and negative log-likelihood (NLL) from the quantile-interpolated distribution. We emphasize NLL here because likelihood is a \emph{proper-scoring rule} \cite{gneiting2007strictly} for sequence generation, while quantile-loss is only proper for a single prediction. Given the intent to use these models for RL simulators, this distinction is relevant.

\subsection{Network architecture}
\label{sec:networkdesign}
\begin{figure}[H]
    \centering
    \includegraphics[width=0.75\textwidth]{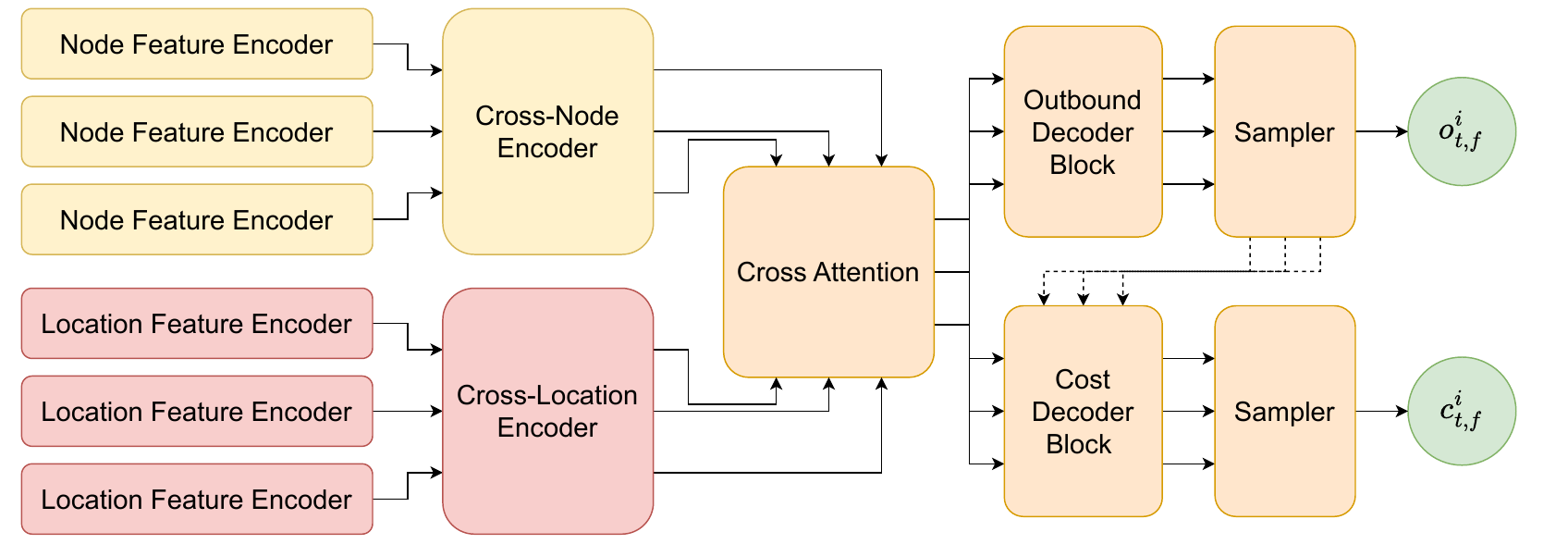}
    \caption{Architecture schema of our proposed model. Dashed Lines between the sampler and Cost Decoder Block represent teacher-forcing during training.}
    \label{fig:network}
\end{figure}

We follow the MQ-Forecaster framework \citep{Wen2017,Eisenach2020}, where historical information are encoded with a stack of dilated convolutions over the time dimension, with the addition customized design to account for sharing of information between different warehouses for aligning warehouse-level data with location-level data. The architecture starts by using a wavenet encoder on two parallel streams of data (glanceview time-series data with location-granularity and warehouse time-series data). This step acts on each node/location locally. The next component involves sharing information between node-level embeddings and location-level embedding via either a bi-drectional RNN or a Transformer. These are chosen such that the model can flexibly adapt to new warehouses. In \cite{ahn2024gnnbasedprobabilisticsupplyinventory}, \cite{mengjin_2024} the authors utilize graph networks to match supply and demand at different granularities. Similarly, \cite{yiling2024} utilizes the cross-attention mechanism to let demand streams attend to each other, resulting in information exchange. In the same spirit, we utilize a cross-attention layer to join the two embedding streams from nodes and locations, allowing for the supply-demand matching we aim to capture. Finally, these embeddings are decoded via MLPs to representations (quantiles and logits) in order to sample outbound. The Outbound is provided as an input for a decoder for a cost head.  In training, we use teacher forcing, providing the cost decoder the actual outbound rather than a sample from the model. See Figure~\ref{fig:network} for a network schematic.

\section{An off-policy backtesting oracle}\label{sec:backtest-outbound}
We describe a methodology to implement an oracle for backtesting out-of-distribution trajectories. The high level idea is to replay each historical glance view, get its promise from a production system, convert it to an order, conditionally on this promise, and then pass this to system F. With the counterfactual fulfillment plan from F, we can estimate a shipping cost with a model. This whole process will be run sequentially through time -- \emph{not parallelized by product} -- in order for the backtest to capture the effect of the counterfactual inventory placement on ``multi-shipments'', i.e. shipments where two or more items are packed together and their shipment cost is tied to the cost of shipping the box rather than the individual items. If, instead, we ran the simulator in parallel across products, we would lose the ability to model cross-product interactions, since these depend on the joint inventory state of all products. The mechanics of multi-shipments are in fact extremely important, as they drive a large portion of the fulfillment behavior by being so effective at reducing shipping costs. --we give more details in appendix \ref{sec:cost-accounting}--. 

\subsection{Conditional Glance View Conversions}\label{subsec:conditional_conversion}

In our backtesting methodology, accurately modeling how historical glance views convert into orders under new promises is crucial. To achieve this, we introduce a \emph{conditional} conversion model that leverages historical data to inform counterfactual scenarios. This can be viewed as a speed-aware extension of availability correction used in \cite{Madeka2022}, where demand is corrected by projecting as if all customers had seen the item as available. This section elucidates the conditional glance view conversion processs. Independent sampling of glance view conversions can lead to high variance and unrealistic scenarios, especially when multiple related orders influence each other over time. By conditioning on historical outcomes, we ensure that the simulated conversions remain consistent with observed behaviors, which may potentially produce more reliable backtest results.

\paragraph{Conceptual Framework}  
Consider a glance view $v$ that historically resulted in a specific ship option under a given promise. When evaluating a new promise $p$, we aim to determine the probability of each possible ship option while respecting the historical decision-making process. This is achieved by conditioning on the historical promise $H_S(v)$ and the historical outcome $H_O(v)$, effectively reusing the underlying randomness that led to the original conversion.

\paragraph{Cumulative Distribution Function (CDF)}  
For each product $i$ at time $t$ and promise $p$, we define the cumulative conversion rate over ship options $\mathcal{O}$ as:
\[
\bar{R}_{t}^{i,p,o} = \sum_{o' \leq o} \hat{R}_{t}^{i,p,o'}
\]
where $\hat{R}_{t}^{i,p,o}$ is the estimated probability that a glance view for product $i$ at time $t$ under promise $p$ converts to ship option $o$. Thus, $\bar{R}_{t}^{i,p,o}$ denotes the probability of conversion at speed $o$ or faster. As conversion rates increase with promise, $\bar{R}_{t}^{i,p,o}$ can be assumed to be monotone increasing as promise speed increases.

This cumulative conversion rate allows us to interpret the conversion process as follows: Imagine that the historical order outcome was generated by drawing a uniform random variable $U \in [0,1]$ and selecting ship option $o$ if
\[
\bar{R}_{t}^{i,p,o-1} < U \leq \bar{R}_{t}^{i,p,o}
\]
where $\bar{R}_{t}^{i,p,o-1}$ is the cumulative conversion rate just faster than ship option $o$.

\paragraph{Conditioning on Historical Outcomes}  
Given a historical glance view with promise \( H_S(v) \) and outcome \( H_O(v) \), we can infer that the underlying \( U \) must have fallen within a specific interval that led to the observed outcome. Specifically, if \( H_O(v) = o_{\text{hist}} \), then:

\[
\bar{R}_{t}^{i,H_S(v),o_{\text{hist}}-1} < U \leq \bar{R}_{t}^{i,H_S(v),o_{\text{hist}}}
\]

By conditioning on this interval, we restrict \( U \) to lie within \( \left( \bar{R}_{t}^{i,H_S(v),o_{\text{hist}}-1}, \bar{R}_{t}^{i,H_S(v),o_{\text{hist}}} \right] \). This ensures that when evaluating a new promise \( p \), the same \( U \) is used, maintaining consistency in the conversion process. Because of the ordering, an increase in the promise can only increase the probability of converting at faster speeds and so if promise increases (or stays the same) and an order converted historically, this sampling process will result in the order converting again, though potentially at a faster ship option.

\paragraph{Conditional Probability Calculation}  
Given the constrained \( U \) from the historical outcome, we compute the conditional probability of each ship option \( o \) under a new promise \( p \). The conditional probability is defined as:
{\small
\begin{align*}\label{eq:conditional_conversion}
    \Pr[o \mid p, H_S(v), H_O(v)] = \nonumber \left(\frac{\min\left(\bar{R}_{t}^{i,p,o}, \bar{R}_{t}^{i,H_S(v), H_O(v)}\right) - \max\left(\bar{R}_{t}^{i,p,o-1}, \bar{R}_{t}^{i,H_S(v), H_O(v)-1}\right)}{\hat{R}_{t}^{i,H_S(v), H_O(v)}}\right)^{+}
\end{align*}
}
For an illustrative example of the conditional glanceview conversion process, see Appendix~\ref{sec:appendix_cond_conv}. This glanceview conversion is integrated into an outbound oracle that loops through glanceviews, converts them, calls system F and tracks inventory which is detailed in Appendix~\ref{sec:oracle:app}.

\section{Empirical Results}\label{sec:experiments}

\begin{table}
\centering
\caption{Validation quantile loss (q), negative log-likelihood (nll), along with discrete cross-entropy (ce). L is for the final combined validation loss. Closest-node Baseline(Appendix~\ref{apd:closestnode}) represents our modeling of a greedy closest-node baseline. Transformer and RNN represent using transformer blocks or a Bidirectional RNN for the Cross-X-encoders in Figure~\ref{fig:network}, while sales indicates whether a feature for sales is included. Note that these are ``diagnostic'' models, since sales cannot be used in an RL dynamics model, as they are endogenous.}
\vspace{9pt}
\scriptsize
\begin{tabular}{lccccccccc c}
\toprule
 & \multicolumn{4}{c}{Cost} & \multicolumn{5}{c}{Outbound} & \multirow{2}{*}{L} \\
\cmidrule(lr){2-5} \cmidrule(lr){6-10}
 & q10 & q50 & q90 & nll & q10 & q50 & q90 & nll & ce \\
\midrule
Closest node              & -    & -     & -     & -    & 0.614 & 0.671 & 0.700 & 15.1  & 8.15 &  - \\
RNN                     & 0.052 & 0.147 & 0.086 & 1.442 & 0.050 & 0.144 & 0.081 & 2.638 & 0.754 & 4.394 \\
Transformer             & 0.052 & 0.148 & 0.087 & 1.442 & 0.050 & 0.145 & 0.082 & 2.634 & 0.758 & 4.355 \\
RNN w sales          & 0.052 & 0.145 & 0.084 & 1.429 & 0.052 & 0.150 & 0.082 & 2.653 & 0.759 & 4.274 \\
Transformer w sales  & 0.052 & 0.144 & 0.083 & 1.412 & 0.049 & 0.142 & 0.078 & 2.621 & 0.752 & 4.247 \\
\bottomrule
\end{tabular}
\label{tab:quantile}
\end{table}

In this section, we evaluate different model architectures with respect to their accuracy and calibration. For accuracy, we use quantile loss benchmarks and compare models against a closest-node baseline, which fulfills each order from the nearest node with available inventory. We selected closest-node as it is commonly accepted as a good humanly interpretable heuristic and used in a number of internal applications. Calibration metrics, instead, can be evaluated independently based on sampling performance. We assess the ability of different cross-node encoder architectures—Bidirectional RNN and Transformer blocks—to predict outbound location and cost. Additionally, we explore whether including a feature for sales (i.e., actual conversions) improves model performance. While this “leaky” sales information is not usable in RL environments due to its endogeneity, it diagnoses the impact of model having to perform the glance view conversion versus not having to do that. \\

All models are trained on a dataset of 50k randomly sampled products spanning July 2022 to December 2023 and validated on the same products between January and June 2024. The loss function is a linear combination of quantile loss, negative log-likelihood, and cross-entropy (as detailed in Section~\ref{sec:lossfunction}). Table~\ref{tab:quantile} shows validation results across cost and outbound prediction tasks. We find that the choice of cross-node encoder (RNN vs. Transformer) has minimal effect on predictive performance. The addition of sales information leads to small improvements in cost prediction, while having no impact on outbound accuracy. This is possibly because sales precisely localize historic demand, which informs cost based fulfillment decisions. To evaluate calibration, we generate 128 samples for 5k validation products per model and aggregate predictions at both national and regional levels. Table~\ref{tab:sample} reports multiplicative calibration slopes for p10, p50, and p90 quantiles, along with OLS slopes and the Continuous Ranked Probability Score (CRPS). Transformer-based models yield the best calibration, especially at the regional level. RNN also performs similarly. 
Instead, the closest-node baselines fails to capture the correct behavior at the regional level. We discuss more of the shortcomings of the baseline model in appendix \ref{apd:closestnode}. Finally, in Figure~\ref{fig:calib_t}, we show that our model remains well-calibrated over time in both the continuous (quantile) and discrete (outbound location) settings. The discrete calibration curves indicate that predicted probabilities align closely with empirical frequencies. \\

We conclude that the candidate architectures are well suited for integration within a simulator to learn inventory control policies. They provide accurate and well-calibrated predictions of outbound dynamics and associated costs over time—a component in defining the reward function of the control policy-. The proposed Oracle backtesting framework enables validation on out-of-distribution counterfactual trajectories, adding an important layer of safety. Moreover, outputs from the Oracle can be used to finetune the models on these trajectories, improving their generalization capabilities. Overall, we view this as a promising component of a framework for learning and validating inventory control policies that operates at a granular regional level.

\begin{table}
\centering
\caption{Samples are taken from the model and aggregated at a national and regional level. Multiplicative calibration is done for empirical quantiles and mean of outbound resulting in calibration slopes (p10, p50, p90), OLS slope, and CRPS for each model at Total and Regional granularity.}
\vspace{9pt}
\scriptsize
\begin{tabular}{llccccc}
\toprule
Model & Aggregation & p10 slope & p50 slope & p90 slope & OLS slope & CRPS \\
\midrule
Closest node      & Total    & 0.75 & 0.94 & 1.11 & 0.91 & 48.51 \\
RNN             & Total    & 0.90 & 0.98 & 1.04 & 1.07 & 34.85 \\
Transformer     & Total    & 0.94 & 0.94 & 0.93 & 0.94 & 20.60 \\
Closest node      & Regional & 0.13 & 0.42 & 1.30 & 0.28 & 26.90 \\
RNN             & Regional & 0.91 & 0.98 & 1.00 & 0.96 &  6.25 \\
Transformer     & Regional & 0.96 & 0.96 & 0.96 & 0.96 &  4.11 \\
\bottomrule
\end{tabular}
\label{tab:sample}
\end{table}

\begin{figure}
    \centering
    \begin{subfigure}[b]{0.46\textwidth}
        \centering
        \includegraphics[width=\textwidth]{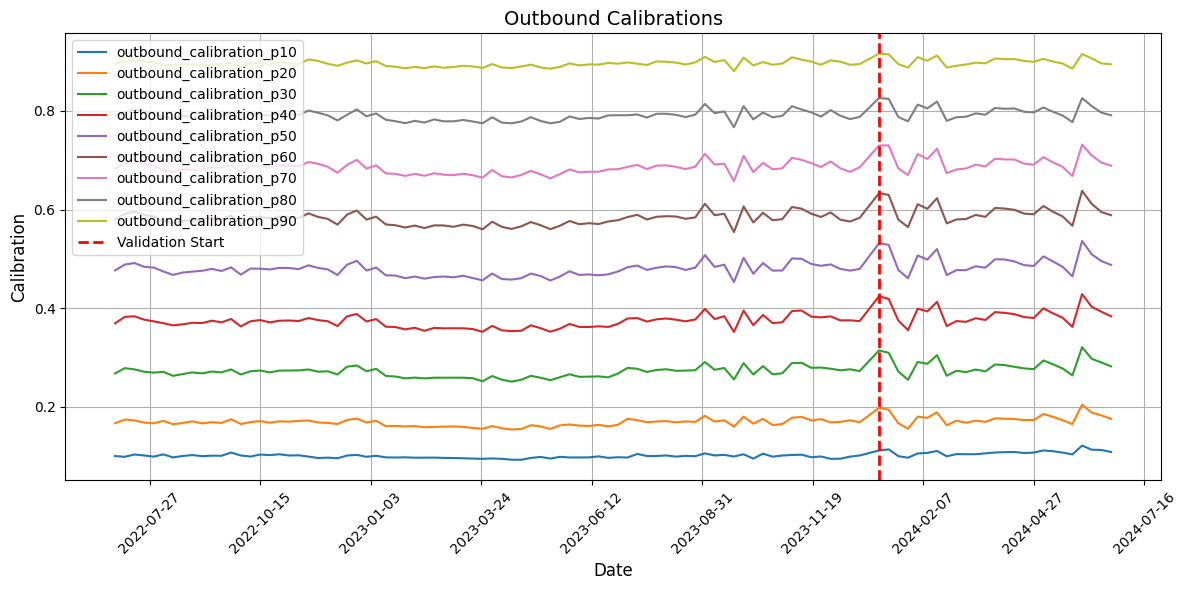}
    \end{subfigure}
    \hfill
    \begin{subfigure}[b]{0.46\textwidth}
        \centering
        \includegraphics[width=\textwidth]{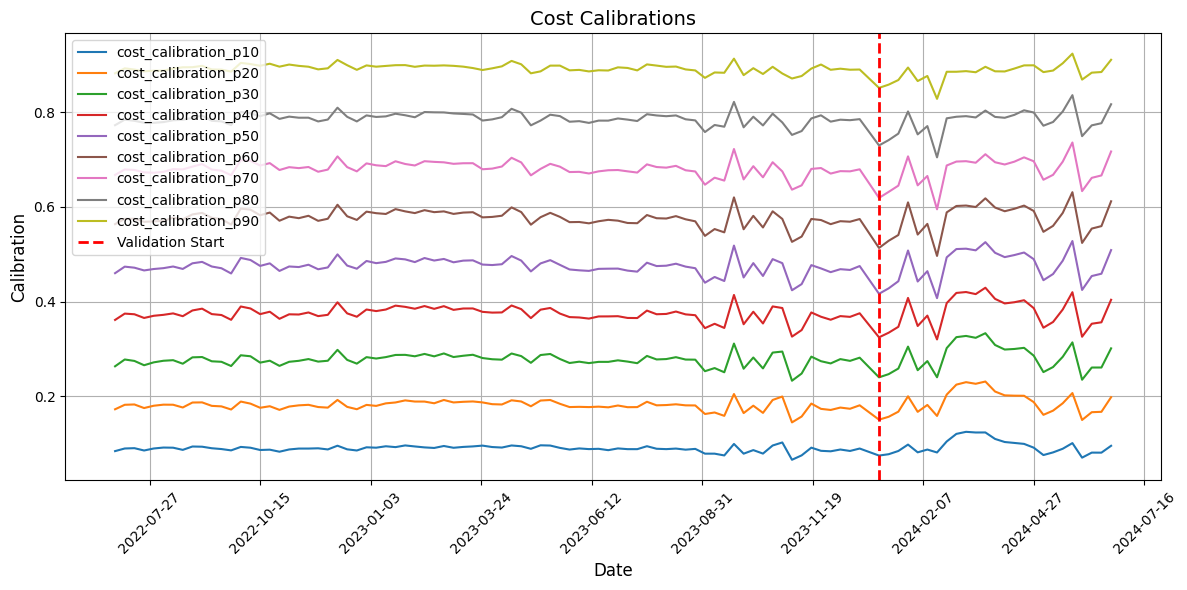}

    \end{subfigure}
    \hfill
    \begin{subfigure}[b]{0.55\textwidth}
        \centering
        \includegraphics[width=\textwidth]{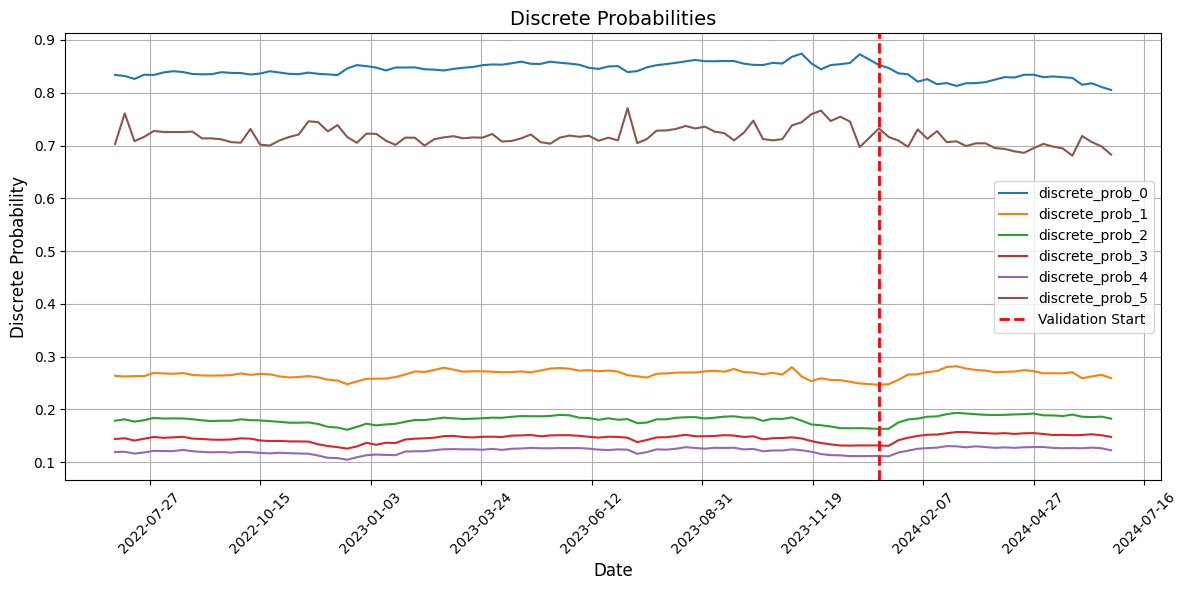}
    \end{subfigure}
    \caption{Calibrations for quantiles and the discrete predictions.  Discrete calibrations represent probability of $k$ given true outbound is $k$. Final probability $5$ represents probability $>= 5$.}
    \label{fig:calib_t}
\end{figure}

\nocite{geraci2021midquantileregressiondiscreteresponses}
\nocite{santos2015bayesianquantileregressionanalysis}
%%%%%%%%%%%%%%%%%%%%%%%%%%%%%%%%%%%%%%%%%%%%%%%%%%%%%%%%%%%%%%%%%%%%%%%
%%%%%%%%% BIBLIOGRAPHY
%%%%%%%%%%%%%%%%%%%%%%%%%%%%%%%%%%%%%%%%%%%%%%%%%%%%%%%%%%%%%%%%%%%%%%%
\clearpage
\bibliographystyle{plain}
% \bibliography{ref}

\begin{thebibliography}{10}

\bibitem{clark1960optimal}
A.~J. Clark and H.~Scarf.
\newblock Optimal policies for a multi-echelon inventory problem.
\newblock {\em Management Science}, 6(4):475--490, 1960.

\bibitem{Eisenach2020}
Carson Eisenach, Yagna Patel, and Dhruv Madeka.
\newblock {MQTransformer: Multi-Horizon Forecasts with Context Dependent and Feedback-Aware Attention}, 2020.

\bibitem{geraci2021midquantileregressiondiscreteresponses}
Marco Geraci and Alessio Farcomeni.
\newblock Mid-quantile regression for discrete responses, 2021.

\bibitem{gneiting2007strictly}
Tilmann Gneiting and Adrian~E. Raftery.
\newblock Strictly proper scoring rules, prediction, and estimation.
\newblock {\em Journal of the American Statistical Association}, 102(477):359--378, 2007.

\bibitem{ahn2024gnnbasedprobabilisticsupplyinventory}
Hyung il~Ahn, Young~Chol Song, Santiago Olivar, Hershel Mehta, and Naveen Tewari.
\newblock Gnn-based probabilistic supply and inventory predictions in supply chain networks, 2024.

\bibitem{mengjin_2024}
Mengjin Liu, Yuxin Zuo, Yang Luo, Daiqiang Wu, Peng Zhen, Jiecheng Guo, and Xiaofeng Gao.
\newblock Weather-conditioned multi-graph network for ride-hailing demand forecasting.
\newblock In {\em Service-Oriented Computing: 22nd International Conference, ICSOC 2024, Tunis, Tunisia, December 3–6, 2024, Proceedings, Part II}, page 341–356, Berlin, Heidelberg, 2024. Springer-Verlag.

\bibitem{Madeka2022}
Dhruv Madeka, Kari Torkkola, Carson Eisenach, Anna Luo, and Dean Foster.
\newblock {Deep Inventory Management}, 2022.

\bibitem{Quenneville-Belair2023}
Vincent Quenneville-Belair, Malcolm Wolff, Brady Willhelme, Dhruv Madeka, and Dean Foster.
\newblock Distribution-free multi-horizon forecasting and vending system.
\newblock In {\em KDD 2023 International Workshop on Mining and Learning from Time Series (MileTS)}, 2023.

\bibitem{santos2015bayesianquantileregressionanalysis}
Bruno Santos and Heleno Bolfarine.
\newblock Bayesian quantile regression analysis for continuous data with a discrete component at zero, 2015.

\bibitem{Wang_2027}
Zirui Wang and Tianying Wang.
\newblock A semiparametric quantile single-index model for zero-inflated and overdispersed outcomes.
\newblock {\em Statistica Sinica}, 2027.

\bibitem{Wen2017}
Ruofeng Wen, Kari Torkkola, Balakrishnan Narayanaswamy, and Dhruv Madeka.
\newblock {A multi-horizon quantile recurrent forecaster}.
\newblock In {\em NIPS Time Series Workshop}, 2017.

\bibitem{yiling2024}
Yiling Wu, Yingping Zhao, Xinfeng Zhang, and Yaowei Wang.
\newblock Spatial–temporal correlation learning for traffic demand prediction.
\newblock {\em IEEE Transactions on Intelligent Transportation Systems}, PP:1--14, 11 2024.

\end{thebibliography}

\newpage
\appendix

\section{Model Details}

\subsection{Features used}

\begin{table}[ht]
\centering
\caption{Node Features}
\vspace{9pt}
\begin{tabular}{ll}
\toprule
\textbf{Feature Name} & \textbf{Description} \\
\midrule
\texttt{outbound}            & Fulfilled demand from this node. \\
\texttt{shipping\_cost}     & Cost or shipping metric for this node. \\
\texttt{node\_location}       & GPS coordinates for this node.       \\
\texttt{available\_inv}      & Current available inventory at node.  \\
\texttt{is\_warehouse\_active} & Binary indicator if node is active.  \\
\bottomrule
\end{tabular}
\end{table}

\begin{table}[ht]
\centering
\caption{Location Features}
\vspace{9pt}
\begin{tabular}{ll}
\toprule
\textbf{Feature Name} & \textbf{Description } \\
\midrule
\texttt{total\_gv} & \makecell[l]{Aggregate glanceviews (webpage visits) \\ for this location.}\\
\texttt{zip\_location}  & GPS coordinates for this location. \\
\bottomrule
\end{tabular}
\end{table}

\begin{table}[ht]
\centering
\caption{Holiday Features}
\vspace{9pt}
\begin{tabular}{ll}
\toprule
\textbf{Feature Name} & \textbf{Description} \\
\midrule
\texttt{distance\_to\_event\_*}    & \makecell[l]{distance measure in days to yearly \\ seasonal events.} \\
\bottomrule
\end{tabular}
\end{table}

\subsection{Loss and optimization details}
The loss is configured as a fixed linear combination of cost nll, cost quantile loss, cross entropy on the discrete outbound predictions, nll on the outbound for the quantile part of the distribution and quantile loss for the outbound.  The weights are $[0, 4, 2, 0.3, 6]$, which were chose to approximately normalize each loss to scale close to $1$ during training, though this is quite arbitrary. Understanding how these losses impact a downstream metric is an interesting area for improvement.

Models are trained for $200-300$ epochs or until validation loss seems steadily increasing, which occured with transformer models. Losses typically converged much quicker, reaching near final loss within 20 epochs.

Models were trained with DDP on 8 40 GB A100 GPUs, allowing for batch sizes on the order of 160 for the models using transformer encoders and 256 for models using the RNN encodes which required less memory.

\subsection{Model hyperparameters}

\textbf{Output Tensors:}
\begin{itemize}
    \item Outbound distribution: 6 discrete logits plus 9 quantiles. 
    \item Cost distribution: continuous quantiles.
\end{itemize}

\textbf{Key Hyperparameters:}
\begin{itemize}
  \item \(\mathtt{hidden\_size}=64\), \(\mathtt{attention\_heads}=8\).
  \item \(\mathtt{rnn\_layers}=2\), or \(\mathtt{transformer\_layers}=2\) 
  \item \(\mathtt{mlp\_depth}=3\), \(\mathtt{dropout}=0.1\).
  \item \(\mathtt{atrous\_rates}=[1,\,2,\,4]\) (dilations in CNN).
  \item \(\mathtt{quantiles} = [0.1,\,0.2,\,0.3,\,0.4,\,0.5,\,0.6,\,0.7,\,0.8,\,0.9]\).
  \item \(\mathtt{num\_outbound\_logits}=6\) (6 discrete demand buckets so outbound $>=5$ is predicted using quantiles).
\end{itemize}

\subsection{Architecture Outline}

\paragraph{(1) Temporal CNN Encoders}
Both \(\mathtt{node\_t}\) and \(\mathtt{location\_t}\) are first concatenated (per time step) with the \(\mathtt{dist\_t}\) features, then each passes through a stack of \textbf{3 dilated convolutions} (WaveNet-style). 
\begin{itemize}
  \item Each convolution has: 
    \(\text{kernel\_size}=2,\) 
    \(\text{out\_channels}=\mathtt{hidden\_size}=64,\) 
    \(\mathtt{dilation}\in\{1,2,4\}.\)
  \item Each layer is followed by \(\mathtt{ELU}\) and a padding step to preserve sequence length.
\end{itemize}
This produces node embeddings \(x \in \mathbb{R}^{B\times T\times N\times 64}\) and location embeddings \(z \in \mathbb{R}^{B\times T\times L\times 64}\).

\paragraph{(2) Channel RNN Encoding}
Since \(\mathtt{transformer\_layers}=0\), we use a bidirectional RNN on the channel dimension (\(N\) or \(L\)):
\begin{itemize}
  \item \(\text{RNN input\_size}=64\), 
  \(\text{hidden\_size}=64/2=32\) per direction,
  \(\text{num\_layers}=2\).
  \item The node embeddings (\(B\times T\times N\times 64\)) are reshaped to \((B\times T)\cdot N\) mini-batches, passed through the RNN, then reshaped back. Location embeddings similarly.
\end{itemize}
We obtain updated node embeddings \(x_r\) and location embeddings \(z_r\), both of shape \((B\times T\times \{\!N\!\!\text{ or }\!L\!\}\times 64)\).

\paragraph{(3) Cross-Attention (Nodes \(\leftarrow\) Locations)}
A multi-head cross-attention module (\(\mathtt{attention\_heads}=8\)) allows each node to attend over location embeddings:
\[
   \text{Att}(x_r,z_r) \;\in\; \mathbb{R}^{B\times T\times N\times 64}.
\]
This becomes the “outbound head.”

\paragraph{(4) Outbound MLP Decoder}
A depth-3 MLP (with \(\mathtt{in\_features}=64,\ \mathtt{out\_features}=\mathtt{num\_outbound\_logits}+Q=10+9=19\)) produces:
\[
   \underbrace{o\_\text{logits}}_{(10\ \text{channels})},\quad
   \underbrace{o\_\text{quantiles}}_{(9\ \text{channels})}.
\]
The logits model discrete demand buckets; the quantile channels (via softplus + cumsum) ensure monotonic quantile outputs.

\paragraph{(5) Cost Quantile Transformer}
We map \((x_r,\,z_r)\) to cost quantiles similarly:
\begin{itemize}
  \item Optionally concatenate the outbound draw (if available) to \(x_r\).
  \item A second cross-attention with location embeddings.
  \item A depth-3 MLP outputs cost quantiles, shape \((B\times T\times N\times 9)\).
\end{itemize}

\paragraph{Summary}
Using the above 5-step pipeline, our model creates distributional predictions of both outbound demand (discrete + quantile) and cost (quantile). The default dimensioning follows:
\begin{itemize}[leftmargin=*,noitemsep]
    \item \textbf{CNN Layers:} 3 dilated convs (\(k=2\), dilation = 1,2,4), each output size = 64.
    \item \textbf{RNN:} Bidirectional, 2 layers, hidden size = 64 total (32 per direction).
    \item \textbf{Attention:} 8 heads.
    \item \textbf{MLPs:} 3-layer fully connected, from 64 up to final channel count (19 for outbound, 9 for cost).
\end{itemize}

\section{Customer Interest Conversion Example}\label{sec:appendix_cond_conv} 
To concretize the conditional conversion mechanism, consider the following example involving a single product and a single time period.

\begin{itemize}
    \item \textbf{Ship Option Cateogries}: \( \mathcal{O} = \{\text{1d}, \text{2d}, \text{3d+}, \text{NoOrder}\} \), indexed as \( o \in \{0, 1, 2, 3\} \).
    \item \textbf{Promise Categories}: \( \mathcal{S} = \{\text{1d}, \text{2d}, \text{3d+}, \text{Out-of-Stock}\},  \).
    \item \textbf{Historical Data} for product \( i \) at time \( t \) under historical promise \( p_h = H_S(v) \):
\end{itemize}

\begin{table}[H]
    \centering
    \footnotesize
    \caption{Historical Conversion Probabilities for Product \( i \) at Time \( t \) under Promise \( p_h =\text{2d} \)}
    \vspace{9pt}
    \label{tab:historical_prob}
    \begin{tabular}{cccc}
        \toprule
        \textbf{Ship Option} & \textbf{Probability} & \textbf{CDF} & \textbf{U Range} \\
        \midrule
        1d        & 0.00  & 0.00  & $[0.00,\ 0.00)$ \\
        2d        & 0.20  & 0.20  & $[0.00,\ 0.20)$ \\
        3d+        & 0.10  & 0.30  & $[0.20,\ 0.30)$ \\
        NoOrder  & 0.70  & 1.00  & $[0.30,\ 1.00]$ \\
        \bottomrule
    \end{tabular}
\end{table}

Suppose the historical outcome was \( H_O(v) = \text{2d} \), which implies that the underlying \( U \) fell within the range \( [0.0, 0.2) \).

\textbf{Backtest Scenario:}
\begin{itemize}
    \item \textbf{New Promise}: \( p_b = \text{1d} \) (improved promise).
    \item \textbf{Conversion Probabilities under \( p_b \)}:
\end{itemize}

\begin{table}[H]
    \centering
    \footnotesize
    \caption{Conversion Probabilities for Product \( i \) at Time \( t \) under New Promise \( p_b \)}
    \vspace{9pt}
    \label{tab:new_prob}
    \begin{tabular}{cccc}
        \toprule
        \textbf{Ship Option} & \textbf{Probability} & \textbf{CDF} & \textbf{U Range} \\
        \midrule
        1d       & 0.15 & 0.15 & $[0.00,\ 0.15)$ \\
        2d       & 0.15 & 0.30 & $[0.15,\ 0.30)$ \\
        3d+      & 0.10 & 0.40 & $[0.30,\ 0.40)$ \\
        NoOrder  & 0.40 & 1.00 & $[0.40,\ 1.00]$ \\
        \bottomrule
    \end{tabular}
\end{table}

\textbf{Conditional Probability Calculation:}

Given that the historical \( U \in [0.0, 0.2) \), we examine how this interval overlaps with the new CDF under \( p_b \) to determine the conditional probabilities for each ship option.

\begin{table}[H]
    \centering
    \footnotesize
    \caption{Overlap of Historical \( U \) with New Promise \( p_b \) CDF}
    \vspace{9pt}
    \label{tab:overlap}
    \begin{tabular}{ccccc}
        \toprule
        \textbf{Ship Option} & \makecell{\textbf{New CDF} \\ \textbf{Range}} & \makecell{\textbf{Overlap with} \\\( [0.0, 0.2) \)} & \makecell{\textbf{Overlap} \\ \textbf{Length}} & \makecell{\textbf{Conditional} \\\textbf{Probability}} \\
        \midrule
        1d       & $[0.00,\ 0.15)$ & $[0.00,\ 0.15)$ & 0.15 & $\frac{0.15}{0.20} = 0.75$ \\
        2d       & $[0.30,\ 0.40)$ & $[0.15,\ 0.20)$ & 0.05 & $\frac{0.05}{0.20} = 0.25$ \\
        3d+      & $[0.40,\ 0.50)$ & None            & 0.00 & 0.00 \\
        NoOrder  & $[0.50,\ 1.00]$ & None            & 0.00 & 0.00 \\
        \bottomrule
    \end{tabular}
\end{table}

\textbf{Summary of Conditional Conversion:}

\begin{table}[H]
    \centering
    \footnotesize
    \caption{Conditional Conversion Probabilities under New Promise \( p_b \)}
    \vspace{9pt}
    \label{tab:conditional_prob}
    \begin{tabular}{cc}
        \toprule
        \textbf{Ship Option} & \textbf{Conditional Probability} \\
        \midrule
        1d       & 0.75 \\
        2d       & 0.25 \\
        3d+      & 0.00 \\
        NoOrder  & 0.00 \\
        \bottomrule
    \end{tabular}
\end{table}

\paragraph{Interpretation of Results}  
From Tables \ref{tab:overlap} and \ref{tab:conditional_prob}, we derive the following conditional probabilities:

\begin{itemize}
    \item \textbf{NoOrder}: The overlap is zero, resulting in a conditional probability of \( 0.00 \). This means that, under the new promise \( p_b \), a historical \( U \) that previously led to an order cannot result in a ``NoOrder'' outcome under the faster promise.
    \item \textbf{1d}: The overlap length of \( 0.15 \) divided by the original interval length of \( 0.20 \) yields a conditional probability of  \( 75\% \). Therefore, there is an \( 75\% \) chance that the ship option becomes ``1d'' under the new promise.
    \item \textbf{2d}: The overlap length of \( 0.05 \) divided by the original interval length of \( 0.20 \) yields a conditional probability of approximately \( 25\% \). Thus, there is a \( 25\% \) chance that the ship option remains ``2d'' under the new promise.
    \item \textbf{3d+}: The overlap is zero, resulting in a conditional probability of \( 0.00 \). This indicates that, under the new promise \( p_b \), there is no probability of selecting the ``3d+'' ship option based on the historical \( U \).
\end{itemize}

This conditional conversion ensures that the historical relationship between promises and outcomes is maintained while adapting to new promises, thereby producing realistic and coherent order outcomes in the backtest scenario.

\section{Oracle}\label{sec:oracle:app}
\subsection{Outbound estimation - production systems}
We will estimate outbound need to define the production systems we need to call. The Promise system can be viewed as a function that maps an inventory configuration for an product to a promise for all $|\cZ|$ regions of the form $F_{pr} :\RR_{\geq 0}^{|\cF|} \times \cA  \rightarrow \cS^{|\cZ|}$. The second system that we need to call to implement an outbound oracle is F, which can be viewed as a stateful system that ingests an inventory configuration $I \in \RR_{\geq 0}^{|\cF|}$ , customer region $ z \in\cZ$ and order and eventually produces a warehouse for the order to be shipped from.

\subsection{Ship cost estimation}
Neither of the two production systems described above will give us a ship cost estimate (the other component of the oracle). For our purposes, we require the ability to sample from the shipment-level distributions. We have an estimated distribution for each time, product, node and region combination, and we denote these as $\hat{P}^{i,f,z}_{t}$. See \ref{sec:estimating-shipment-costs} for more detail.

\subsection{Putting it all together}

\ref{alg:oracle} describes an oracle that takes an inventory state and sequence of glance views and returns a vector of outbound quantities for each product. We use Python-style array indexing for ease of exposition. Observe that this oracle requires a number of service calls linear in the number of orders (not glance views).  Note that final accounting of an order is delayed from when it is placed, as is done in practice. This is purposeful to accurately emulate multi-shipments, though the control flow will be more complicated in real implementation.

\begin{algorithm}
  \caption{Outbound Oracle}
  \label{alg:oracle}
  \begin{algorithmic}
    \Input $I^i \in \RR_{\geq 0}^{|\cF|}$, $\cG = (v_s)_{s \in \ZZ_{+}}$, $\{\hat{R}^{i, p,o}\}$, $\{ \hat{P}^{f,z} \}$
    \State $O^{i} \gets (0,...,0) \forall i$ \texttt{// Total outbound units per node}
    \State $C^{i} \gets (0,...,0) \forall i$ \texttt{// Total ship cost per node}
    \State $P \gets F_{pr}(I)$ 
    \For {$v \in \cG$}
        \State $z' \gets H_Z(v)$ \texttt{// Get region for current glance view}
        \State $i' \gets H_A(v)$ \texttt{// Get product for current glance view}
        \State $p' \gets P[z']$ \texttt{// Get promise for current glance view}
        \State $o' \sim \Pr[\cdot | p', H_S(v), H_O(v)]$ \texttt{// Sample conversion to order}% via \eqref{eq:conditional_conversion}}
        \If{ $o' ~~!=~ \texttt{NoOrder}$}
            \State \texttt{// Order is placed, F adds to fulfillment set}
            \State $(I^{i'},z',o') \rightarrow \text{F}$ \texttt{//Send F orders}
        \EndIf
        
        \If{\text{F.updated()}} \texttt{// F assigns previous orders}
        \State $f'', i'', o'' \gets \text{F}$ \texttt{// Get region for current glance view}
        \State $c' \sim \hat{P}^{i'', f'',z''}$ \texttt{// Sample ship cost}
        \State $O^{i''}[f'] \gets O^{i''}[f''] + 1$
        \State $I^{i''}[f'] \gets O^{i''}[f''] - 1$
        \State $C^{i''}[f'] \gets C^{i''}[f''] + c'$
        \State $P \gets F_{pr}(I, i)$  
        \EndIf
    \EndFor
    \Output $O^{i}$, $C^{i}$ for all products $i$
  \end{algorithmic}
\end{algorithm}

\section{Shipping costs}
\label{sec:cost-accounting}

Here we address two important facts regarding the cost of fulfillment: its definition and the implications of such definition for the \textit{evaluation} of an inventory placement policy. In general, the fulfillment cost can be decomposed as a sum of the costs incurred for packing and delivering a box to a customer. A desirable decomposition method is one that it is representative of the total costs incurred at a company level, but also causally attributes certain costs at a box-unit level. The guarantee on the total costs allows our simulated evaluation to be consistent with what we would see in a real environment, while the box-unit level attribution allow us to attribute a granular reward based on the outcome of inventory placement actions -i.e. the resulting fulfillment trajectory- that can be used to learn an inventory placement policy. An example of this attribution is to start from a  ``ground truth'' cost of shipping a box of a given weight and size, and split the shipment cost of the box's final delivery to the customer across units within that box, according to some criteria -i.e. their volume and weight-. 

The latter example highlights the implication of such attribution method: shipment costs may not be independent across ordered products, as for example in this case packing two items together significantly modifies the fulfillment cost of both. Changing the shipment modality thus implies the counterfactual cost \textbf{has} to be computed conditionally to all shipped units that \textit{could have} been packed together, across all products. So while our simulator operates at a product level, any ground-truth backtesting methodology \textbf{must} account for counterfactual cross-product effects to correctly \textit{evaluate} a policy. \\ 

\subsection{Estimating the Shipping Cost}
\label{sec:estimating-shipment-costs}
We here provide a way to estimate the shipping cost of a box of items, based on some generic characteristics of a shipment. This estimate allows us to recreate a cost within the simulator described in \ref{sec:oracle:app}. In fact, for the purpose of \textit{evaluating} the results of an inventory control policy in a simulator, we obviously desire that the simulated costs are representative of the ones that we'd incur in reality.

Since boxes have clearly attributed costs of shipments, by design, the cost accounting represents the total cost incurred for shipments. As such, if we can recreate an unbiased estimate in our simulator, then the cost that our simulator will output should also be representative of the total bill incurred. Naturally, the real costs depend on characteristics of the shipment (characteristics of the package, distance traveled and so on), which are influenced by the inventory control process. We can make our forecast depends on such characteristics, which allows us to forecast costs accurately even when off-policy.

As an exercise, we regress (OLS) the shipping cost of a box $sc^{\mathcal{B}_k}$ over a set of features that reasonably depend from the inventory control process and show the results in Table~\ref{tab:olsboxcost}. For proprietary reasons, we mask features and all the regression coefficients. All of them are statistically significant with $p < 0.001$, and the high R-squared is indicative of a fairly well performing model, regardless of its simplicity.

\begin{table}[h!]
\begin{center}
\caption{OLS regression of the ``ground-truth'' shipping cost per box $sc^{\cB_k}$ in respect of physical characteristics of the shipment.}
\footnotesize
\begin{tabular}{lclc}
\toprule
\textbf{Dep. Variable:}    &        $sc^{\cB_k}$         & \textbf{  R-squared:         } &     0.509   \\
\textbf{Model:}            &       OLS        & \textbf{  Adj. R-squared:    } &     0.509   \\
\textbf{Method:}           &  Least Squares   & \textbf{  F-statistic:       } &     3993.   \\
\textbf{No. Observations:} &       50000      & \textbf{  Prob (F-statistic):} &     0.00    \\
\textbf{Df Residuals:}     &       49986      & \textbf{  Log-Likelihood:    } &   -97744.   \\
\textbf{Df Model:}         &          13      & \textbf{  AIC:               } & 1.955e+05   \\
\textbf{Covariance Type:}  &    nonrobust     & \textbf{  BIC:               } & 1.956e+05   \\
\bottomrule
\end{tabular}
\label{tab:olsboxcost}
\end{center}
\end{table}

%%%%%%
%% Auxiliary models 
%%%%%%

\section{Baseline Model: Closest-node}
\label{apd:closestnode}
In this section we discuss our choice of Closest-node as a baseline, present its detailed results and discuss its performance. Closest-node describes an algorithm that, given a sequence of customer orders, sequentially assigns the closest node with inventory to fulfill the next order, until all orders have been fulfilled or inventory has ran out. Closest-node is anecdotally accepted to be a good heuristic as well as being human interpretable.

\paragraph{Algorithm:} We here describe our implementation of this algorithm, specifically how it generates a vector (over warehouses) of outbound quantities for a given product and time. For every product and time interval, we call for the regional vectors of glance views $g^i \in \ZZ^{|\cZ|}$, national conversion rate $c^i \in [0,1]$, inventory in each warehouse $I^i \in \ZZ_{\geq 0}^{|\cF|}$ and a distance function between regions and warehouses $d : \cZ \times \cF \rightarrow \RR_{\geq 0} $.

\begin{algorithm}[h]
  \caption{Closest-node}
  \label{alg:closest-fc}
  \begin{algorithmic}
    \Input $I^i \in \ZZ_{\geq 0}^{|\cF|}$, $g^i \in \ZZ_{\geq 0}^{|\cZ|}$, $c^i \in [0,1]$, $d$

    \State $O^{i} \gets 0$ \quad \texttt{// Initiate outbound vector at 0}
    \For {$z \in \cZ$}
        \State $s^i_{z} \sim \mathrm{Binomial}(\theta = c^i, n = g^i_{z})$ \quad \texttt{// Sample conversion to order via Binomials}
    \EndFor

    \While {$\sum_z s^i_{z} > 0 \wedge \sum_f I^i_{f} > 0$}
        \State $p^i \gets s^i / \sum_z s^i_z$ \quad \texttt{// Get share of demand per region}
        \State $z' \sim \mathrm{Multinomial}(\vec{\theta} = p^i, n = 1)$ \quad \texttt{// Sample region from Multinomial}
        \State $\cF' \gets \{f \in \cF: I^{i}_f > 0\}$ \quad \texttt{// Find set of warehouses with inventory}
        \For {$f \in \cF'$}
            \State $\delta^{\,i}_{f} \gets d(z', f)$ \texttt{// Get vector of distances}
        \EndFor
        \State $j' \sim \mathrm{Multinomial}(\vec{\theta} = \mathrm{softmax}(-\delta^{\,i}), n = 1)$ \quad \texttt{// Sample closest warehouse}
        \State $O^{i}_{j'} \gets O^{i}_{j'} + 1$ \quad \texttt{// Update outbound}
        \State $I^{i}_{j'} \gets I^{i}_{j'} - 1$ \quad \texttt{// Update inventory}
        \State $s^{i}_{z'} \gets s^{i}_{z'} - 1$ \quad \texttt{// Update orders}
    \EndWhile 
    
    \Output $O^{i}$
  \end{algorithmic}
\end{algorithm}

Products, times and samples are considered independent of each other; in practice we thus vectorized the algorithm across these dimensions with the appropriate considerations. 
Note also that this algorithm relaxes the notion of closest node with a softmax sampling, which allows inventory to be drawn, with probability, from other nodes in the proximity. This is general enough to represent the Closest-node notion: as we tune the multiplier hyperparameter, the softmax sampling acts more and more deterministically, degenerating in an argmin function and thus coinciding with the closest node notion. \\

\begin{figure}
    \centering
    \includegraphics[width=0.46\textwidth]{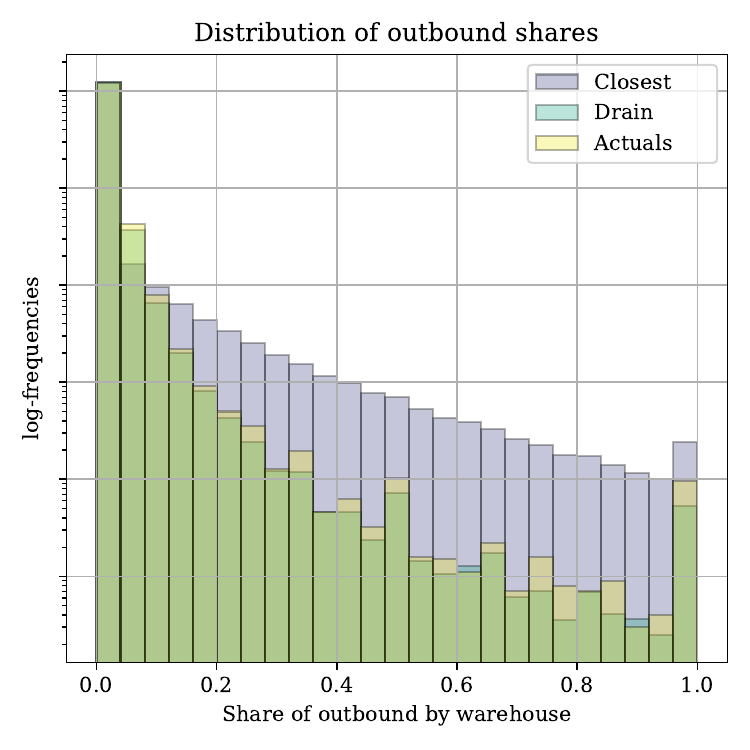}
    \caption{The log-frequencies of shares of outbound per warehouse. A share of 1 implies all outbound for a product-week is performed by a single warehouse. It is visible from the closest node distribution that it tends to concentrate outbound in few nodes, while the NN based Drain model tends to resemble the distribution of actual outbound.}
    \label{fig:fancy3d}
\end{figure}

\paragraph{Experiment and Results:} We ran the algorithm with deterministic sampling\footnote{Future work will include ablations on the temperature parameter to see if it improves performance.} -i.e. actual closest node- to obtain 128 samples for each product-time combination, and used the empirical distributions of the samples as predictions for the results in table \ref{tab:quantile}. Model performance in all metrics is very poor compared to neural-net based models. The reason is that the model outputs distributions of outbound units which are overly concentrated, i.e. all outbound is performed by a few warehouses, which are usually close to large cities where clusters of demand are present. Figure \ref{fig:fancy3d} is an exemplification of this fact. 

\paragraph{Discussion and future work:} We attribute the underperforming results of this model to a multitude of reasons. For starter, the data generation process concentrates all customer demand to have equal GPS coordinates in each region, thus resulting in one node always taking up all the demand for a particular region. More in general, closest node is a simplistic algorithm that does not take into account a multitude of factors that determines the F outbound decision, such as transportation costs, shipments costs, capacity, customer shipment consolidation and others. For these reasons we believe that closest node needs to improve and consider other variables before reaching the sophistication necessary to mimic F behavior. We're currently in the process of expanding the research on these baseline models, as there is clear value in good models whose properties and behavior are human interpretable.

\section{A model for Glance View conversion}
\label{sec:gvconversion}
We present here a model to sample the conversion of a glance view into an order. At a high level, the model assumes that glance views are independent and have conversion probabilities that are exogenous given a product, time and region combination. If we have estimates of the conversion probabilities, we can replay historic glance views with counterfactual promises to simulate orders.

We now define our choice of $\cO$ and $\cS$. Promises are grouped into three types: one-day (1d), two-day (2d), three or more days (3d+) and out-of-stock (-). Similarly, we group ship options into four types: one-day (1d), two-day (2d), three or more days (3d+) and no order (-). Thus we have $\cS = \{1d, 2d, 3d+\}$ and $\cO = \{1d, 2d, 3d+, - \}$.

We model each glance view as an independent sample from a multinomial distribution where promise $j \in \cS$ has probability $p_{j,k}$ of converting to SO $k \in \cO$. The model can be represented as a $|\cS| \times |\cO|$ matrix as in \ref{tab:conversionmodel}. Note that there are never ship-options available that are faster than the promise shown to the customer, and thus those conversion probabilities are always 0. Similarly, when the item is out-of-stock, the probability of no order is 1. 

\begin{table}[H]
    \centering
    \caption{A simple model for glance views conversions based on probabilities.}
    \begin{tabular}{c c||c c c c}
       & & \multicolumn{3}{c}{Ship Option} \\
       & & No-Order & 1d & 2d & 3d+  \\ \hline \hline
    \multirow{3}{*}{\rotatebox[origin=c]{90}{Speed}} 
     & Out-of-Stock & 1 & 0 & 0 & 0 \\
     & 1d  & $p_{1d,-}$ & $p_{1d,1d}$ & $p_{1d,2d}$ & $p_{1d,3d+}$  \\
     & 2d  & $p_{2d,-}$ & 0 & $p_{2d,2d}$ & $p_{2d,3d+}$ \\
     & 3d+ & $p_{3d+,-}$ & 0 & 0 & $p_{3d+,3d+}$ 
    \end{tabular}

    \label{tab:conversionmodel}
\end{table}

We estimate the elements of the matrix using historical data.

\end{document}